\documentclass{article}
\usepackage{spconf,amsmath,graphicx,hyperref}
\usepackage{xeCJK} % XeLaTeXで日本語を有効にする
\usepackage{amsmath, amssymb}
\usepackage{booktabs}
\usepackage{subcaption}

\title{Leveraging Label Proportion Prior \\ for Class-Imbalanced Semi-Supervised Learning}
\name{Kohki Akiba, Shinnosuke Matsuo, Shota Harada, Ryoma Bise}
\address{Kyushu University, Fukuoka, Japan}
\begin{document}
\ninept
\maketitle
\begin{abstract}
Semi-supervised learning (SSL) often suffers under class imbalance, where pseudo-labeling amplifies majority bias and suppresses minority performance. 
We address this issue with a lightweight framework that, to our knowledge, is the first to introduce Proportion Loss from learning from label proportions (LLP) into SSL as a regularization term. 
Proportion Loss aligns model predictions with the global class distribution, mitigating bias across both majority and minority classes. 
To further stabilize training, we formulate a stochastic variant that accounts for fluctuations in mini-batch composition. 
Experiments on the Long-tailed CIFAR-10 benchmark show that integrating Proportion Loss into FixMatch and ReMixMatch consistently improves performance over the baselines across imbalance severities and label ratios, and achieves competitive or superior results compared to existing CISSL methods, particularly under scarce-label conditions.

\end{abstract}

\section{Introduction}
\label{sec:intro}

% In many real-world scenarios, the distribution of classes is highly imbalanced, with certain categories represented by only a small number of examples. This imbalance poses a fundamental challenge for semi-supervised learning (SSL), where pseudo-labeling is widely used to exploit unlabeled data. Once the underlying classifier develops a bias toward majority classes, pseudo-labels generated from it propagate the same bias, resulting in further suppression of minority-class performance. Consequently, SSL that performs strongly on balanced benchmarks often suffers a substantial degradation in accuracy when applied to imbalanced data.

Semi-supervised learning (SSL) has emerged as a powerful paradigm~\cite{lee2013pseudo, grandvalet2004semi, verma2019interpolation, laine2017temporal, tarvainen2017mean, miyato2018virtual, berthelot2019mixmatch,  sohn2020fixmatch} to leverage large amounts of unlabeled data alongside a small set of labeled samples. 
This is often achieved via pseudo-labeling~\cite{lee2013pseudo}, where a classifier assigns labels to unlabeled data to augment the training process.

However, read-world data is often highly imbalanced, with certain categories represented by only a small number of examples~\cite{buda2018systematic, cui2019class}. 
This imbalance poses a fundamental challenge for SSL: once the underlying classifier develops a bias toward majority classes, the pseudo-labels generated from it propagate the same bias, resulting in further suppression of minority-class performance~\cite{kim2020darp, wei2021crest}. 
Consequently, SSL methods that perform strongly on balanced benchmarks often suffer substantial degradation in accuracy when applied to imbalanced data.

In practice, even a small number of labeled samples can provide a rough but informative estimate of the overall class proportions. Such information has been actively explored in the field of learning from label proportions (LLP)~\cite{ardehaly2017co}, where supervision is given in terms of class proportions for groups of samples rather than labels for individual instances. Motivated by these findings, we consider that incorporating class proportion information into SSL can mitigate the adverse effects of imbalance by guiding the learning process toward consistency with the global distribution.

% -------------------------------------------------------------
\begin{figure}[t]
    \centering
    \includegraphics[width=.8\linewidth]{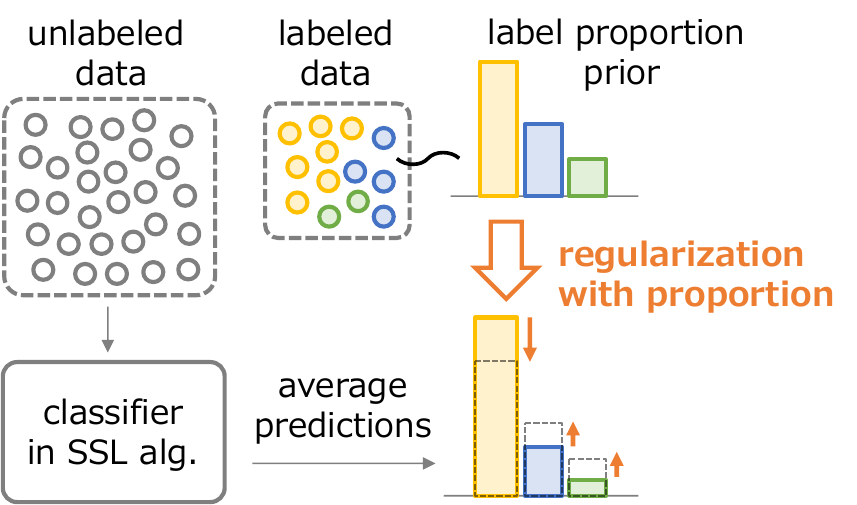}
    \caption{Our approach: Regularization with the label proportion prior for semi-supervised learning (SSL).}
    \label{fig:overview}
\end{figure}
% -------------------------------------------------------------

To this end, we propose a lightweight framework that incorporates Proportion Loss, originally studied in LLP~\cite{ardehaly2017co}, into SSL as a regularization term. Proportion Loss encourages the model predictions to align with the global class distribution, thereby mitigating bias by adjusting both over-represented and under-represented classes.
To our knowledge, this is the first work to bring the idea of label proportions from LLP into the SSL setting. 
The method is conceptually simple yet highly effective, and importantly, it can be seamlessly integrated into a wide range of existing SSL algorithms without requiring architectural modifications.

A further challenge arises from mini-batch training: since batches contain only a small subset of samples, the observed class proportions fluctuate around the global distribution, and naïvely enforcing them can lead to overfitting. We draw inspiration from large-bag LLP to model expected batch composition in a probabilistic manner. This yields a stochastic variant of Proportion Loss that reduces overfitting to spurious batch fluctuations and helps stabilize training under severe imbalance.

We validate our approach on the Long-tailed CIFAR-10 benchmark~\cite{cui2019class}. 
Incorporating Proportion Loss into FixMatch~\cite{sohn2020fixmatch} and ReMixMatch~\cite{berthelot2020remixmatch} improves performance across imbalance severities and label ratios. 
Our method is particularly effective with scarce labels, and remains competitive with imbalance-aware methods when more labels are available. 
It consistently outperforms the baselines, while analysis shows that Proportion Loss mitigates minor-class underestimation, leading to better pseudo-label selection and accuracy.

Our contributions are summarized as follows:
\begin{itemize}
\item We introduce Proportion Loss, originally studied in LLP, into the SSL setting. This simple yet effective regularization aligns mini-batch predictions with the global class distribution, mitigating the bias amplification caused by pseudo-labeling. The method is broadly applicable and can be seamlessly integrated into existing SSL algorithms. 
\item We further develop a stochastic variant of Proportion Loss, inspired by large-bag LLP~\cite{kubo2024theoretical}, that models expected batch composition with a multivariate hypergeometric distribution. This enhances robustness against mini-batch fluctuations under severe imbalance.
\item We validate both variants on the Long-tailed CIFAR-10 benchmark, showing consistent improvements over the baselines. Our method is especially strong under scarce-label conditions, outperforming existing works.
\end{itemize}

\section{Related work}
\label{sec:related}
\noindent
{\bf Semi-supervised learning (SSL):}
Standard SSL methods leverage techniques such as entropy minimization~\cite{grandvalet2004semi} and Mixup regularization~\cite{berthelot2019mixmatch,verma2019interpolation} to effectively utilize unlabeled data.
Another key approach is consistency regularization~\cite{laine2017temporal, tarvainen2017mean, miyato2018virtual}, which enforces similar predictions for perturbed versions of an input and is widely unified with pseudo-labeling. For example,ReMixMatch~\cite{berthelot2020remixmatch} combines Distribution Alignment (DA) with various auxiliary tasks, whereas FixMatch~\cite{sohn2020fixmatch} simplifies this by enforcing consistency between hard pseudo-labels from weakly augmented inputs and their strongly augmented versions.

\noindent
{\bf Class-imbalanced semi-supervised learning (CISSL):}
Compared to standard SSL, relatively few studies have explored CISSL.
DARP~\cite{kim2020darp} refines biased pseudo-labels by solving a convex optimization problem that aligns predictions with estimated class distributions.
CReST~\cite{wei2021crest}, a more recent self-training method, progressively mitigates class imbalance by favoring pseudo-labeled samples predicted as minority classes.

\noindent
{\bf Learning from label proportion:}
LLP~\cite{ardehaly2017co,tsai2020learning,matsuo2023learning,asanomi2023mixbag} addresses scenarios where only the label proportions of a group of instances (bag) are provided instead of instance-level labels. A common approach is to minimize the discrepancy between predicted and given proportions, often referred to as the {\it Proportion Loss}. This aggregate-level supervision enables learning without requiring fully annotated data.
LLP has been studied as an independent problem setting and applied to a variety of domains where precise annotation is costly or infeasible. For example, in digital pathology~\cite{matsuo2024lplp}, and endoscopy~\cite{okuo2025weakly}. Recent studies have further extended LLP to more challenging scenarios, including large bags, where theoretical analyses of proportion label perturbation have been conducted~\cite{kubo2024theoretical}, and partial proportion settings, where only subset-level proportions are available~\cite{matsuo2024lplp}.  

Inspired by these findings, we take a different perspective: instead of treating LLP as a standalone paradigm, we incorporate label proportion information into SSL for the first time, explicitly aligning pseudo-labels with the global class distribution and thereby mitigating bias toward majority classes.
Crucially, while existing techniques like DA~\cite{berthelot2020remixmatch} rescales individual predictions to match a target distribution, our method introduces Proportion Loss as an explicit loss-level regularization. By modeling mini-batch composition probabilistically, our method enforces distribution consistency directly within the learning objective—an approach complementary to DA, as evidenced by the performance gains when combined with ReMixMatch.

\section{Proportion-Regularized Semi-Supervised Learning}
\label{sec:method}
\subsection{Problem Setting}
We consider an $L$-class classification problem under a class-imbalanced semi-supervised learning (SSL) scenario. The labeled dataset is denoted as $X=\{(x_i,y_i):i=1,\dots,N\}$, where $x_i \in \mathbb{R}^d$ is the $i$-th labeled sample and $y_i \in \{1,\dots,L\}$ is its class label. Let $N_l$ be the number of labeled samples from class $l$, satisfying $\sum_{l=1}^L N_l = N$. In addition, we are given an unlabeled dataset $U=\{u_i:i=1,\dots,M\}$ with $u_i \in \mathbb{R}^d$, which is assumed to follow the same class distribution as $X$. Our goal is to learn a classifier $f:\mathbb{R}^d \to \{1,\dots,L\}$ using both $X$ and $U$, under the assumption that the underlying class distribution is imbalanced.

\subsection{Proportion Regularization for SSL}
% -------------------------------------------------------------
\begin{figure}[t]
    \centering
    \includegraphics[width=.99\linewidth]{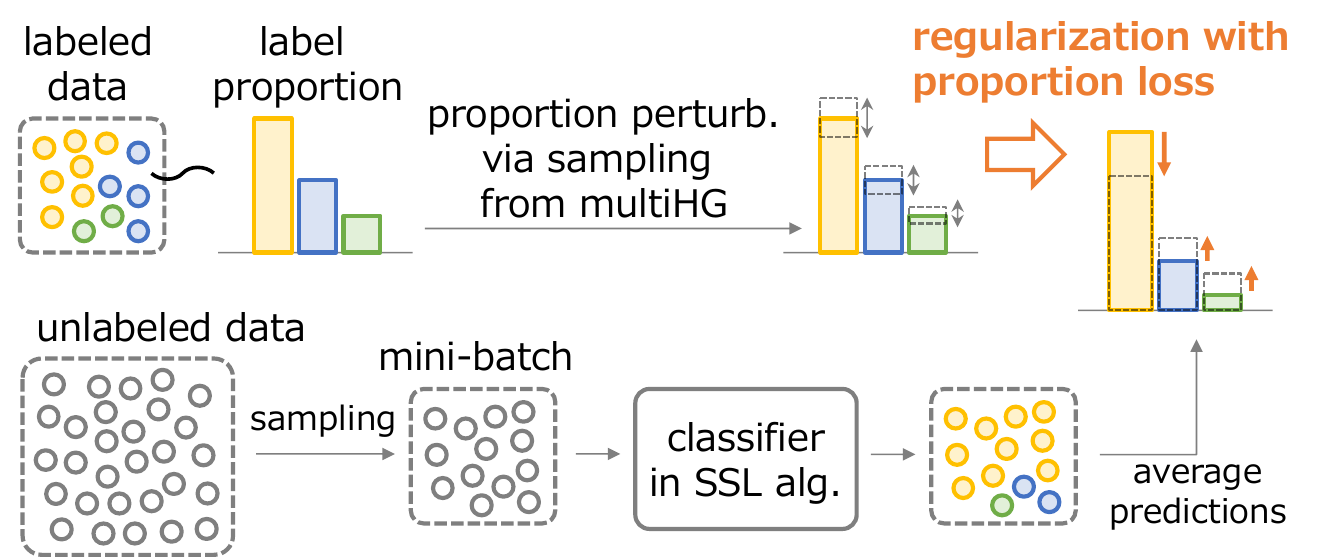}
    % \vspace{-5mm}
    \caption{Overview of the proposed method.}
    \label{fig:model}
\end{figure}
% -------------------------------------------------------------
\subsubsection{Regularization with Proportion Loss}
A key idea of our method is to introduce class proportion information into SSL training. 
We begin by recalling the standard {\it Proportion Loss} widely used in LLP. 
Given a mini-batch $\mathcal{B}=\{u_i\}_{i=1}^{|\mathcal{B}|}$ sampled from the unlabeled set $U$, 
let $\hat{p}_l(\mathcal{B})$ denote the predicted proportion of class $l$, obtained by averaging the model's softmax outputs over the batch. 
Let $\mathbf{q} = (q_1, \dots, q_L)$ denote the estimated global class proportion vector, 
where $q_l = N_l / N$ is the relative frequency of class $l$ in the labeled set $X$ with $N=\sum_{l=1}^L N_l$. 
The Proportion Loss is then defined as:
\begin{equation}
\mathcal{L}_{prop}(\mathcal{B}) = - \sum_{l=1}^L q_l \log \hat{p}_l(\mathcal{B}),
\end{equation}
which encourages the predictions within a batch to be consistent with the global class distribution.

\begin{table*}[th]
\centering
\caption{Test accuracy (\%) on CIFAR-10-LT under different imbalance ratios $\gamma$ and labeled data ratios $\beta$. 
Results are averaged over five runs with different random seeds.}
\label{tab:accuracy_comparison}
\vspace{-2mm}
\begin{tabular}{lcccc}
\toprule
Method & $\gamma=10$, $\beta=2\%$ & $\gamma=20$, $\beta=4\%$ & $\gamma=50$, $\beta=10\%$ & $\gamma=100$, $\beta=20\%$ \\
\midrule
FixMatch & 80.8±0.11 & 78.2±0.14 & 74.5±0.19 & 71.1±0.22 \\
FixMatch + DARP & \underline{81.3±0.10} & \underline{79.1±0.13} & 74.3±0.19 & 71.0±0.23 \\
FixMatch + (CReST+PDA) & 79.6±0.12 & 76.4±0.14 & \textbf{76.6±0.16} & \textbf{73.0±0.19} \\
FixMatch + Ours & \textbf{81.9±0.09} & \textbf{80.4±0.11} & \underline{75.7±0.17} & \underline{72.8±0.20} \\
\midrule
ReMixMatch & 85.5±0.08 & 83.4±0.10 & 79.3±0.14 & 75.5±0.19 \\
ReMixMatch + DARP & \underline{87.5±0.08} & \underline{85.3±0.09} & \textbf{81.5±0.13} & \textbf{77.8±0.17} \\
ReMixMatch + (CReST+PDA) & 86.3±0.09 & 83.9±0.09 & 78.4±0.13 & 75.9±0.16 \\
ReMixMatch + Ours & \textbf{88.1±0.07} & \textbf{85.6±0.09} & \underline{81.2±0.13} & \underline{77.1±0.18} \\
\bottomrule
\end{tabular}
\end{table*}

This proportion constraint can be incorporated into any SSL method as an additional regularization term. 
If the loss of the base SSL method is denoted by $\mathcal{L}_{ssl}$, the overall objective is formulated as:
\begin{equation}
\mathcal{L} = \mathcal{L}_{ssl} + \lambda \mathcal{L}_{prop},
\end{equation}
where $\lambda$ is a hyperparameter that controls the contribution of the Proportion Loss.

\subsubsection{Proportion Perturbation via Hypergeometric Sampling}
Unlike standard LLP, the ground-truth class distribution of a mini-batch sampled from the unlabeled dataset is not directly observable in SSL. 
If we naively use the global class proportion estimated from the labeled set as the supervision for every mini-batch, the supervision inevitably contains noise, especially when the mini-batch size $|B|$ is much smaller than the entire dataset size $N+M$. 
Such fixed noisy supervision can cause the model to overfit, leading the network to converge to biased proportions rather than improving instance-level accuracy~\cite{kubo2024theoretical}.

To mitigate overfitting, we introduce perturbation into the supervision of mini-batch proportions. 
At each iteration $t$, the supervised proportion $\mathbf{q}^{(t)}$ is randomly drawn from a multivariate hypergeometric distribution:
\begin{equation}
\mathbf{q}^{(t)} \sim \text{MultiHG}(M, \mathbf{q}, |B|),
\end{equation}
where $M$ is the total number of unlabeled samples, $\mathbf{q}$ is the estimated global class proportion, and $|B|$ is the mini-batch size.
Intuitively, $\text{MultiHG}(M, \mathbf{q}, |B|)$ models the class proportions obtained when drawing $|B|$ samples without replacement from a population of $M$ items whose class composition follows $\mathbf{q}$. 
The perturbed proportion $\mathbf{q}^{(t)}$ then replaces $\mathbf{q}$ in Eq.~(1) when computing the Proportion Loss, 
thereby adapting the supervision to the stochastic variation of class proportions that arises from random sampling.

This procedure introduces iteration-dependent perturbations that prevent the network from memorizing a fixed noisy proportion. 
Although the Proportion Loss itself does not converge due to the stochastic perturbation, the feature extractor benefits from regularization toward statistically grounded targets. 
As a result, the model reduces majority-class dominance and achieves more reliable instance-level performance.

\section{Experiment}
\label{sec:experiment}
\subsection{Experimental Setup}
\paragraph*{Dataset.}
We evaluate our method on CIFAR-10-LT, a long-tailed variant of CIFAR-10 commonly used in prior studies on imbalanced SSL. 
Following the standard protocol, classes are indexed in descending order of size, with $N_1$ and $N_K$ denoting the largest and smallest classes, respectively. 
The imbalanced training set is constructed by exponentially decreasing the number of samples per class according to an imbalance ratio $\gamma$:
\begin{equation}
    N_k = N_1 \times \gamma^{-\frac{k-1}{K-1}}, \quad \gamma = \frac{N_1}{N_K},
\end{equation}
where $N_k$ is the number of labeled samples in the $k$-th class and $K$ the total number of classes.
We vary $\gamma \in \{10, 20, 50, 100\}$, setting $N_1 \in \{90, 180, 450, 900\}$ accordingly, and also vary the labeled data ratio $\beta \in \{2\%, 4\%, 10\%, 20\%\}$, where $\beta = N/(M+N)$. 
When $\beta$ is small and $\gamma$ large (e.g., $\gamma=100$), minor classes have only a few labeled samples, making evaluation unreliable since all methods fail. 
Thus, we report results only under settings where at least a minimal number of labeled samples (e.g., 9 images when calculated as $4500\beta/\gamma$) are available: $(\gamma,\beta)=(10,2\%), (20,4\%), (50,10\%), (100,20\%)$. 

For validation, 500 images per class are randomly sampled to form a balanced validation set, and final performance is reported on the test set using the checkpoint with the best validation accuracy.

\noindent
{\bf Model architecture.}
Following prior work on class-imbalanced semi-supervised learning (CISSL)~\cite{kim2020darp, wei2021crest}, 
we adopt Wide ResNet-28-2~\cite{zagoruyko2016wrn} as the backbone network for all methods. 
This choice ensures a fair comparison with existing approaches and has been widely used as a standard architecture in SSL benchmarks.

% \paragraph*{Baselines and comparison methods.}
\noindent
{\bf Baselines and comparison methods.}
We used FixMatch~\cite{sohn2020fixmatch} and ReMixMatch~\cite{berthelot2020remixmatch} as baseline SSL methods, 
and evaluated our approach by incorporating the proposed regularization into them. 
We also compared with existing methods for CISSL, namely DARP~\cite{kim2020darp} and CReST~\cite{wei2021crest}, which were also applied to FixMatch and ReMixMatch for fair comparison.

\begin{figure*}[t]
    \centering
    \begin{subfigure}[t]{0.24\linewidth}
        \centering
        \includegraphics[width=\linewidth]{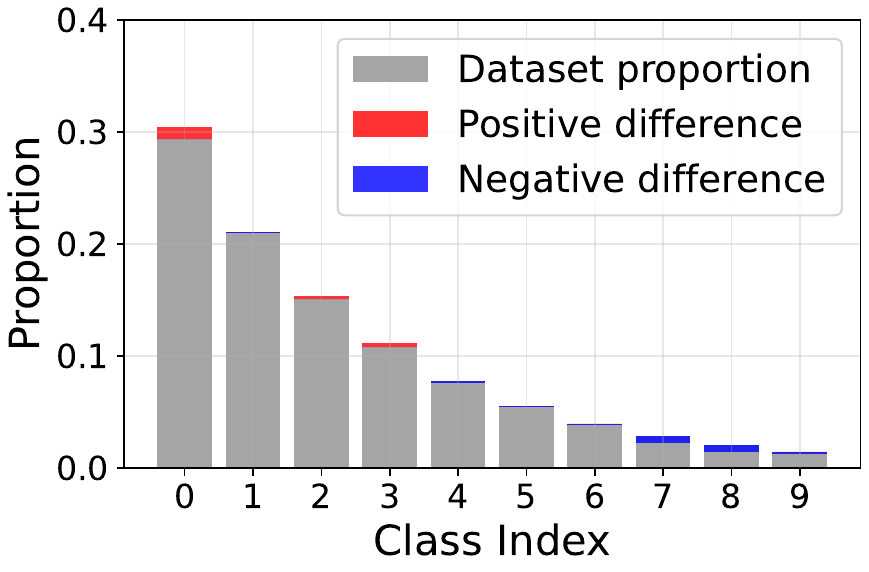}
        \vspace{-5mm}
        \caption{FixMatch}
    \end{subfigure}
    \hfill
    \begin{subfigure}[t]{0.24\linewidth}
        \centering
        \includegraphics[width=\linewidth]{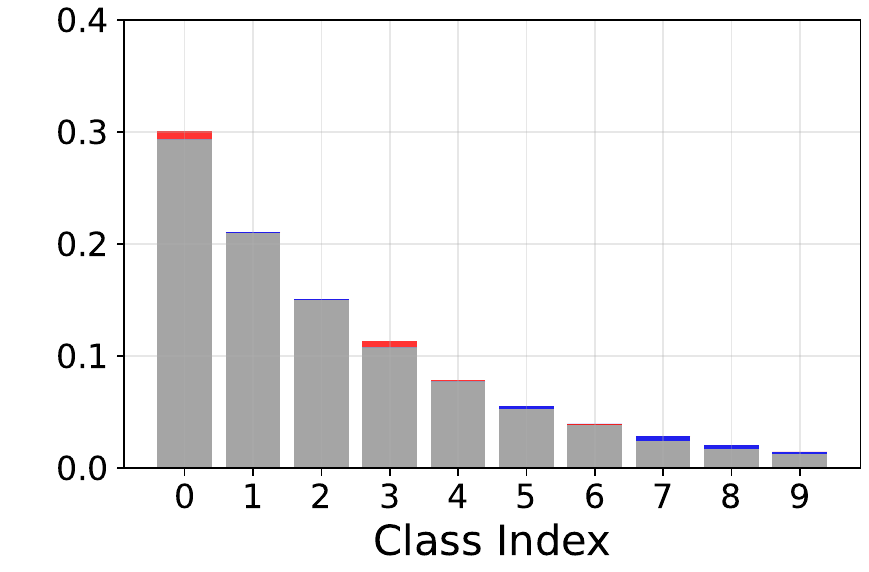}
        \vspace{-5mm}
        \caption{FixMatch + DARP}
    \end{subfigure}
    \hfill
    \begin{subfigure}[t]{0.24\linewidth}
        \centering
        \includegraphics[width=\linewidth]{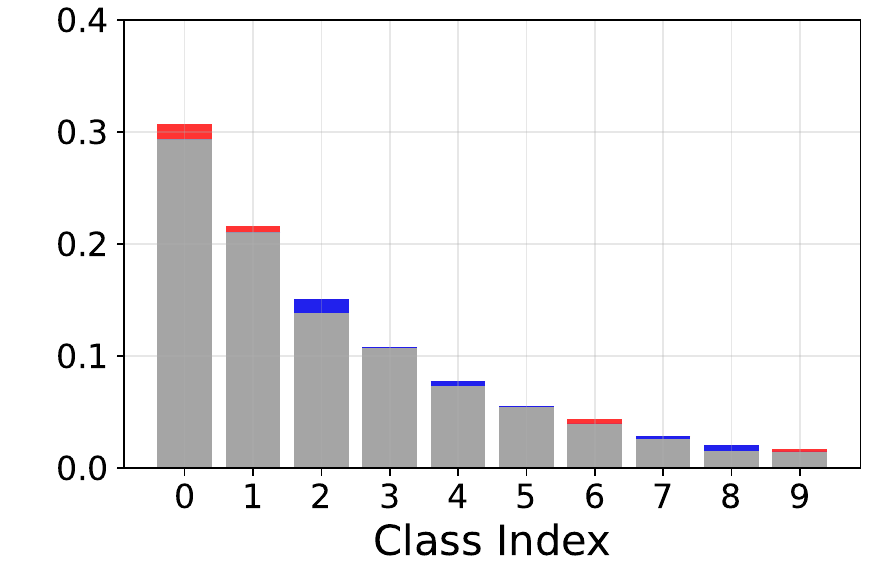}
        \vspace{-5mm}
        \caption{FixMatch + (CReST+PDA)}
    \end{subfigure}
    \hfill
    \begin{subfigure}[t]{0.24\linewidth}
        \centering
        \includegraphics[width=\linewidth]{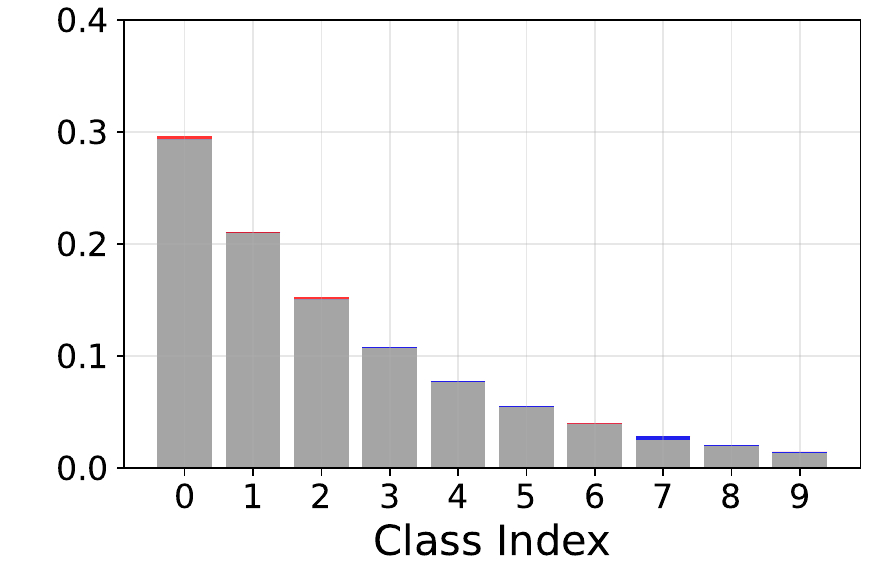}
        \vspace{-5mm}
        \caption{FixMatch + Ours}
    \end{subfigure}
    \vspace{-2mm}
    \caption{Comparison of estimated output proportions after training. Red indicates overestimation and blue indicates underestimation across classes (Class 1 = major, Classes 7–9 = minor).}
    \label{fig:proportion}
\end{figure*}

\begin{figure}[t]
    \centering
    \begin{subfigure}[t]{0.49\linewidth}
        \centering
        \includegraphics[width=\linewidth]{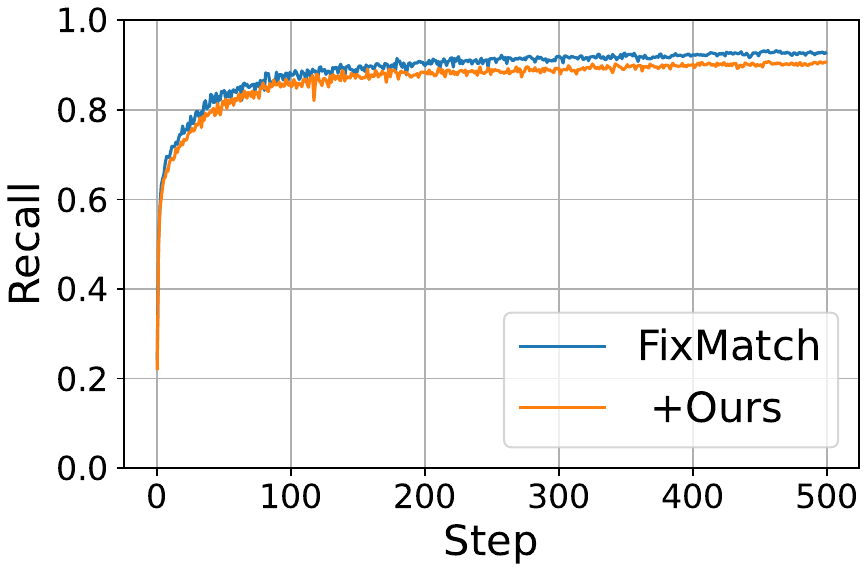}
        \vspace{-5mm}
        \caption{Major class}
    \end{subfigure}
    \hfill
    \begin{subfigure}[t]{0.49\linewidth}
        \centering
        \includegraphics[width=\linewidth]{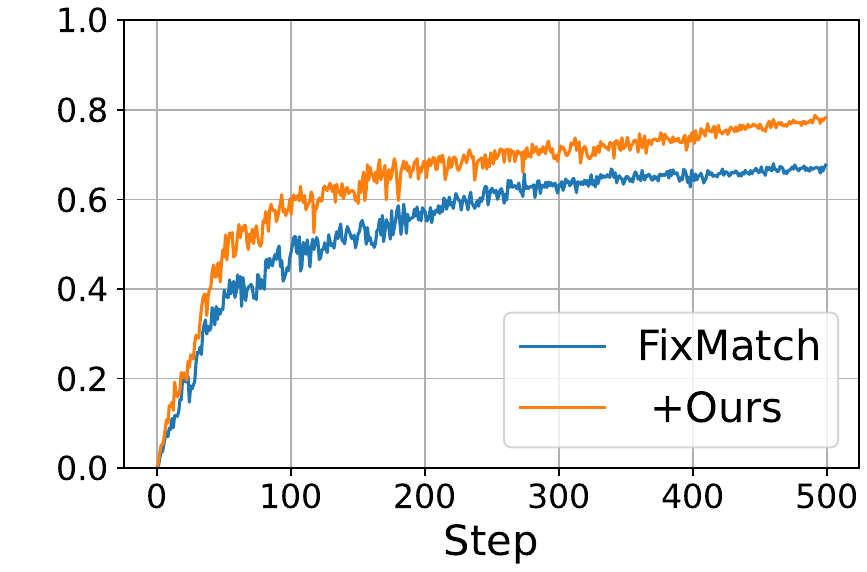}
        \vspace{-5mm}
        \caption{Minor class}
    \end{subfigure}
    \vspace{-2mm}
    \caption{Recall of pseudo-labels for the most major class (left) and most minor class (right) during training.}
    \label{fig:recall}
\end{figure}

% \paragraph*{Training details.}
\noindent
{\bf Training details.}
All models are trained for 500 epochs with 500 iterations per epoch. 
Following the FixMatch setup~\cite{sohn2020fixmatch}, each training iteration uses a batch size of 64 labeled and 448 unlabeled samples (i.e., $\mu = 7$). 
The optimizer is SGD with momentum 0.9 and weight decay $5 \times 10^{-4}$, 
and the initial learning rate is set to 0.03 with a cosine learning rate scheduler~\cite{loshchilov2017sgdr}.

For evaluation, we selected the checkpoint that achieves the best validation accuracy during training. 
The weight $\lambda$ for the Proportion Loss is tuned on the validation set.
Final results are reported as the mean and standard deviation over five independent runs with different random seeds.

\subsection{Comparison of Accuracy}
As shown in Table~\ref{tab:accuracy_comparison}, our method achieves the best performance across both FixMatch and ReMixMatch backbones when the labeled data ratio is small ($\beta=2\%, 4\%$). 
Interestingly, the behavior of existing imbalance-aware methods differs depending on the backbone: with FixMatch, CReST does not provide additional benefit and DARP yields the best result, whereas with ReMixMatch, DARP becomes less effective and CReST provides larger gains. 
Note that across all settings, the number of labeled samples in each minor class is fixed at 9, and smaller $\beta$ values correspond to fewer labeled samples overall. 
As the labeled ratio increases to $10\%$ or $20\%$, existing methods such as DARP and CReST become more effective and sometimes achieve the best accuracy; nevertheless, our method consistently ranks within the top two across all conditions. 
Importantly, it improves upon the baseline FixMatch and ReMixMatch in \emph{all} settings, demonstrating robustness and stability regardless of the backbone, imbalance ratio, or labeled data ratio.

\subsection{Estimated Proportion after Training}
\label{sec:ex_proportion}
To confirm the effectiveness of regularization with the Proportion Loss, we evaluated the output proportion from the unlabeled data for each method: Baseline (FixMatch), DARP, and the proposed method. 
Fig.~\ref{fig:proportion} shows the proportion of the estimated output after training, where red indicates overestimation and blue indicates underestimation.
In the Baseline (FixMatch, Fig.~\ref{fig:proportion} (a)), the estimated proportions deviate from the ground truth. In particular, the major class (Class 1) is overestimated, while the minor classes (Class 7, 8, 9) are underestimated. In DARP and CReST+PDA (Fig.~\ref{fig:proportion} (b) and (c)), the difference in proportion is reduced compared to the Baseline; however, overestimation of the major class and underestimation of the minor classes still remain. Our method further reduces this discrepancy and mitigates the underestimation of minor classes, as shown in Fig.~\ref{fig:proportion} (d). This mitigation positively impacts pseudo-label selection and leads to improved accuracy.

\subsection{Analysis of Pseudo-Label Selection}
As shown in Section~\ref{sec:ex_proportion}, our method alleviates the underestimation of minor classes by regularizing the output proportion of unlabeled data. 
This adjustment provides a more reliable distributional prior, which in turn improves the quality of pseudo-label selection. 
To investigate this effect, we measured the recall of pseudo-labels for major and minor classes during training in both the Baseline (FixMatch) and the proposed method.
Fig.~\ref{fig:recall} shows the recall of pseudo-labels over training steps for the most major class (left) and the most minor class (right). 
Our method significantly improved the recall of the minor class compared to the Baseline, while keeping the recall for the major class at a level comparable to the Baseline.

\section{Conclusion}
\label{sec:conclusion}
% %目的・背景
% In this study, we proposed a method to address CISSL, where the bias amplification caused by pseudo-labeling not only undermines the performance of minority classes but also deteriorates the overall classification accuracy.
% To tackle this issue, we incorporated label-proportion information into SSL training by adapting Proportion Loss from LLP, so that it can counteract the tendency of pseudo-labels to overfit majority classes. We further formulated a stochastic variant based on the multivariate hypergeometric distribution when computing the Proportion Loss. This can mitigate overfitting to incidental mini-batch fluctuations and provide stable improvements across varying degrees of imbalance.

% %結果
% Experiments demonstrated that our method consistently enhances the performance of both FixMatch and ReMixMatch across different imbalance severities and labeled data ratios. In particular, it achieves superior results over both the baselines and existing CISSL methods when the labeled data ratio is low ($\beta=2\%,4\%$).

In this study, we addressed CISSL, where pseudo-labeling amplifies class imbalance and degrades both minority performance and overall accuracy. 
To our knowledge, we are the first to incorporate label-proportion information from LLP into SSL by adapting Proportion Loss as a regularization term, aligning pseudo-labels with the global class distribution and thereby correcting class-level biases. 
We further proposed a stochastic variant based on the multivariate hypergeometric distribution to account for mini-batch fluctuations, which improves stability under severe imbalance. 
Experiments on CIFAR-10-LT demonstrated consistent gains for both FixMatch and ReMixMatch across imbalance severities and label ratios, with clear advantages over the baselines and competitive or superior performance compared to existing CISSL methods, particularly when the labeled data ratio is small ($\beta=2\%,4\%$). 

%limitation
However, several limitations remain. First, due to the nature of our method, its effectiveness may deteriorate when the labeled and unlabeled data follow different distributions. Second, when the mini-batch size of unlabeled data is small, the label proportions cannot be estimated with sufficient accuracy, which may reduce the benefit of proportion regularization. We leave addressing these limitations to future work.

\vspace{2mm}
\noindent
\textbf{{Acknowledgements:}} This work was supported by JSPS KAKENHI Grant Number JP25K22846, ASPIRE Grant Number JPMJAP2403, and JST ACT-X Grant Number JPMJAX23CR.

\clearpage
% -------------------------------------------------------------------------
\bibliographystyle{IEEEbib}
\bibliography{main}

\end{document}